\documentclass[11pt]{article}

\usepackage[preprint]{acl}

\usepackage{times}
\usepackage{latexsym}

\usepackage[T1]{fontenc}

\usepackage[utf8]{inputenc}

\usepackage{microtype}
\usepackage{float}

\usepackage{inconsolata}

\usepackage{graphicx}
\usepackage{amsmath}
\usepackage{amssymb}
\usepackage{amsfonts} 
\usepackage{newtxtext}
\usepackage{newtxmath}
\usepackage{dsfont}
\usepackage{dirtytalk}
\usepackage{soul}
\usepackage{dsfont}
\usepackage[most]{tcolorbox}
\usepackage{xcolor}
\usepackage{hyperref}
\usepackage{fontawesome5} 
\usepackage{booktabs}
\usepackage{arydshln}
\usepackage{multirow}
\usepackage[table]{xcolor}
\usepackage[dvipsnames]{xcolor}
\usepackage[most]{tcolorbox}
\usepackage[T1]{fontenc}
\usepackage{inconsolata} 
\usepackage{listings}
\usepackage{subcaption}
\usepackage{pifont}
\usepackage{caption}
\usepackage{fontawesome5}
\usepackage{enumitem}
\usepackage{needspace}

\DeclareCaptionType{case}[Case][List of Cases]

\tcbuselibrary{skins, breakable}

\definecolor{col_primary}{RGB}{20, 20, 60}   
\definecolor{col_bg_primary}{RGB}{240, 240, 255}     
\definecolor{col_bg_think}{RGB}{255, 230, 255}   
\definecolor{col_frame_think}{RGB}{160, 0, 160}

\definecolor{col_bg_correct}{RGB}{240, 253, 244}
\definecolor{col_frame_correct}{RGB}{22, 163, 74}

\definecolor{col_bg_wrong}{RGB}{254, 242, 242}
\definecolor{col_frame_wrong}{RGB}{220, 38, 38}

\newcommand{\cmark}{\textcolor{green!60!black}{\ding{51}}} 
\newcommand{\xmark}{\textcolor{red!70!black}{\ding{55}}}   

\definecolor{gaincolor}{RGB}{34, 139, 34} 
\definecolor{reductioncolor}{RGB}{178, 34, 34}
\definecolor{basecolor}{gray}{0.25}       

\newcommand{\entry}[3]{%
  \textcolor{basecolor}{#1} $\to$ \textbf{#2} \textcolor{gaincolor}{\fontsize{7}{8}\selectfont{(\textbf{#3})}}%
}
\newcommand{\reduction}[3]{%
  \textcolor{basecolor}{#1} $\to$ \textbf{#2} \textcolor{reductioncolor}{\fontsize{7}{8}\selectfont{(\textbf{#3})}}%
}

\definecolor{rowgray}{gray}{0.96}

\tcbset{
  grpoeqbox/.style={
    enhanced,
    colback=gray!2,
    colframe=gray!45,
    boxrule=0.7pt,
    arc=3pt,
    left=1pt,
    right=3pt,
    top=1pt,
    bottom=1pt,
  }
}

\lstdefinestyle{promptstyle}{
  basicstyle=\ttfamily\small,
  breaklines=true,
  columns=fullflexible,
  numbers=none,
  showstringspaces=false,
}

\newenvironment{questionbox}[1]{
    \tcolorbox[
        enhanced,
        breakable,
        width=\textwidth, 
        boxrule=0pt, leftrule=3pt,
        arc=2pt,
        colback=col_bg_primary,
        colframe=col_primary,
        coltitle=col_primary,
        fonttitle=\bfseries\large,
        title={\faQuestionCircle\ \textbf{#1}},
        attach title to upper,
        after title={\par\vspace{0.5em}\hrule\vspace{0.5em}},
    ]
}{
    \endtcolorbox
}

\newenvironment{thinkingbox}[1]{
    \tcolorbox[
        enhanced, 
        breakable,
        width=\textwidth, 
        boxrule=0pt, leftrule=3pt,
        arc=2pt,
        colback=col_bg_think,
        colframe=col_frame_think,
        coltitle=col_frame_think,
        fonttitle=\bfseries\large,
        title={\faBrain\ \textbf{#1}},
        attach title to upper,
        after title={\par\vspace{0.5em}\hrule\vspace{0.5em}},
    ]
}{
    \endtcolorbox
}

\newtcolorbox{answerboxcorrect}[1][]{
    enhanced,
    width=\textwidth, 
    boxrule=1pt,
    colback=col_bg_correct,
    colframe=col_frame_correct,
    coltitle=col_frame_correct,
    fonttitle=\bfseries,
    title={\faCheckCircle\ \textbf{Correct Output}},
    attach title to upper,
    after title={\par\vspace{0.2em}},
    #1
}

\newtcolorbox{answerboxwrong}[1][]{
    enhanced,
    width=\textwidth, 
    boxrule=1pt,
    colback=col_bg_wrong,
    colframe=col_frame_wrong,
    coltitle=col_frame_wrong,
    fonttitle=\bfseries,
    title={\faTimesCircle\ \textbf{Incorrect Output}},
    attach title to upper,
    after title={\par\vspace{0.2em}},
    #1
}

\title{AMIR-GRPO: Inducing Implicit Preference Signals into GRPO}

\author{Amir Hossein Yari \qquad Fajri Koto \\
        Department of Natural Language Processing, MBZUAI\\
        \small \texttt{\{amir.yari,fajri.koto\}@mbzuai.ac.ae} \\ \faGithub \, \href{https://github.com/AmirHosseinYari2002/AMIR-GRPO} {\textsf{\small https://github.com/AmirHosseinYari2002/AMIR-GRPO}}}

\begin{document}
\maketitle
\begin{abstract}

Reinforcement learning has become the primary paradigm for aligning large language models (LLMs) on complex reasoning tasks, with group relative policy optimization (GRPO) widely used in large-scale post-training. However, GRPO faces structural limitations in reasoning-heavy settings: sequence-level advantage normalization introduces systematic length bias, penalties for low-quality trajectories are diluted, and the scalar objective discards rich pairwise preference information embedded in within-group reward rankings. As a result, valuable supervision from costly rollouts remains underutilized.

We propose AMIR-GRPO, which augments GRPO with an implicit DPO-style contrastive regularizer constructed directly from intra-group reward rankings, requiring no additional annotations. This mechanism amplifies suppression of low-reward trajectories, attenuates response-level length bias, and transforms each rollout group into a denser set of supervision constraints. Across multiple mathematical reasoning benchmarks, AMIR-GRPO consistently outperforms strong GRPO baselines, yields clearer separation between correct and incorrect reasoning chains, and delivers broader coverage gains beyond the subset of instances solved by standard GRPO.
\end{abstract}

\section{Introduction}

Post-training has become the primary mechanism for scaling the reasoning capabilities of large language models (LLMs). Unlike supervised fine-tuning, reinforcement learning (RL) from outcome-level feedback directly optimizes for high-quality reasoning behavior \citep{openai_learning_to_reason_2024, guo2025deepseek}. Within this paradigm, group relative policy optimization (GRPO) \cite{shao2024deepseekmath} has emerged as a popular approach: it samples multiple completions per query and normalizes rewards within each group, producing a low-variance policy-gradient estimator while eliminating the need for a separate value network \citep{shao2024deepseekmath, zheng2025group}.

In parallel, preference-based objectives such as Direct Preference Optimization (DPO) \citep{rafailov2023direct} effectively sharpen decision boundaries by contrasting preferred and rejected responses through pairwise supervision. However, these approaches fundamentally rely on large curated preference datasets, which are expensive to construct—especially for mathematical reasoning, where high-quality labels are scarce.

These paradigms offer complementary strengths: outcome-based reinforcement learning scales through absolute rewards, while contrastive learning efficiently leverages relative quality signals. However, standard GRPO does not fully exploit the ordering it induces within each rollout group. A rollout group of size $G$ induces up to $O(G^2)$ pairwise preference relations, yet standard GRPO collapses this rich signal into only $O(G)$ scalar advantages. We argue that incorporating explicit contrastive objectives into GRPO could unlock these latent signals, improving sample efficiency and strengthening reasoning alignment.

%

Furthermore, specific design choices in GRPO limit its efficacy in shaping reasoning. First, assigning a single trajectory-level advantage uniformly to all tokens introduces a systematic length bias: with positive advantages, shorter answers receive larger per-token updates, implicitly favoring brevity even when extended reasoning is required; with negative advantages, penalties are diluted over long responses, weakening suppression of long, low-reward trajectories \citep{liu2025understanding}. Second, in the sparse-reward regime characteristic of difficult reasoning tasks, group normalization yields weak penalties on incorrect trajectories. When few completions are correct and many are wrong, the negative advantage assigned to errors is diluted, failing to effectively suppress long, incorrect chains of thought. We argue that adding contrastive objectives to GRPO mitigates these issues by shifting optimization from uniform token-level updates to relative completion-level comparisons. Contrastive signals strengthen penalties for low-quality reasoning through pairwise comparisons and fully exploit the ordering within rollout groups—boosting sample efficiency and alignment with reasoning quality.


These observations lead to our primary research question: \textcolor{black!80}{\emph{Can we enrich group-based policy optimization with contrastive, preference-like structure—without requiring explicit pairwise labels while preserving GRPO’s scalability?}} We answer this question affirmatively by augmenting GRPO with an implicit preference objective derived directly from its own rollouts.

We introduce \textit{\textbf{Augmented Merit-guided Implicit-preference Regularization for GRPO}} (AMIR-GRPO), which injects DPO-style contrastive supervision into GRPO while preserving its simplicity and off-policy compatibility. AMIR-GRPO combines GRPO’s scalability in environment-driven rollouts with the boundary-sharpening effect of contrastive preference learning. The core idea is to reinterpret reward orderings within each rollout group as implicit pairwise preferences: for a fixed query, lower-reward completions act as rejected outputs and are explicitly contrasted against higher-reward completions. This turns each low-reward rollout into a first-class negative example, enabling the policy not only to reinforce effective reasoning patterns but also to more directly suppress undesirable ones. Consequently, AMIR-GRPO mitigates response-level length bias by enforcing negative gradients on long, low-reward trajectories, and it amplifies penalties in sparse-reward settings by converting each low-reward completion into multiple contrastive negatives. Empirically, this framework yields less collapsed, more coverage-enhancing reasoning behavior than standard GRPO, together with a sharper decision boundary.

Across multiple mathematical benchmarks, AMIR-GRPO consistently improves reasoning-focused post-training over strong GRPO baselines. Models trained with AMIR-GRPO achieve higher accuracy, exhibit clearer separation between correct and incorrect trajectories, and show reduced response-length bias. Our analysis suggests that standard GRPO mainly reweights the base model’s existing reasoning modes, whereas AMIR-GRPO better preserves diversity while increasing alignment with high-quality chains. Notably, AMIR-GRPO expands effective competence: rather than only redistributing probability over already-solved instances, it solves additional problems that remain unsolved by both the base model and a tuned GRPO model.


\begin{figure*}[t]
    \centering
    \includegraphics[width=\textwidth]{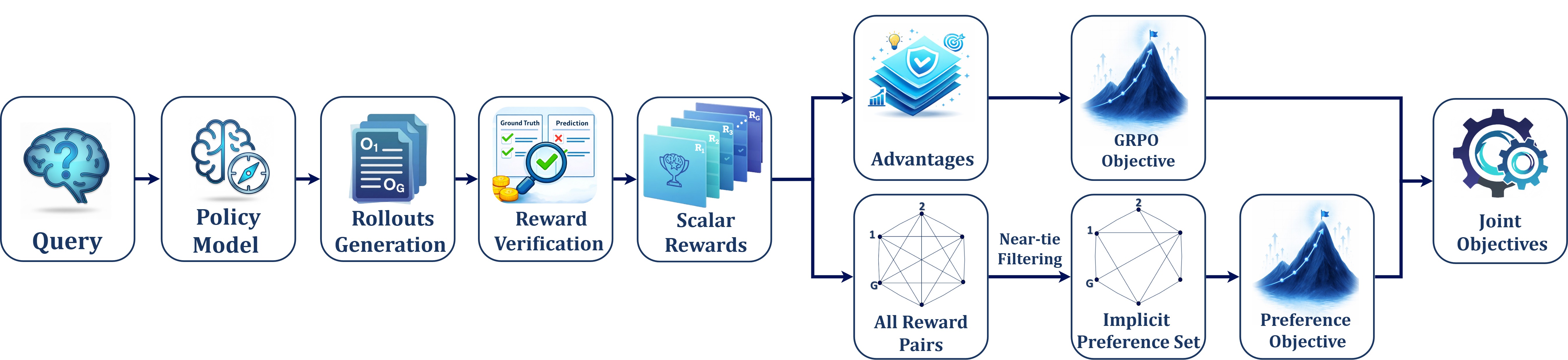}
    \vspace{-0.5cm}
    \caption{End-to-end training pipeline of AMIR-GRPO.}
    \label{fig:pipeline}
\end{figure*}

\section{Methodology}

\subsection{Preliminaries}
\textbf{GRPO} is a PPO-style algorithm that enables value-free policy learning through group-relative credit assignment. For each query $q$, the behavior policy $\pi_{\theta_{\mathrm{old}}}$ samples a group of $G$ candidate responses ${\{o_i\}}_{i=1}^G$ with scalar rewards ${\{r_i\}}_{i=1}^G$. The normalized intra-group advantage for response $i$ is defined as
\begingroup
\setlength{\abovedisplayskip}{0.15cm}
\setlength{\belowdisplayskip}{0.15cm}
\[
    \hat{A}_{i,t}
    =
    \frac{r_i - \operatorname{mean}\!\left(\{r_j\}_{j=1}^G\right)}
         {\operatorname{std}\!\left(\{r_j\}_{j=1}^G\right)}.
\]
which is then shared across all tokens along its trajectory. Using these advantages, GRPO maximizes the clipped surrogate objective in Eq.~\eqref{eq:grpo}. The expectation in $\mathcal{J}_{\mathrm{GRPO}}(\theta)$ is taken over queries $q$ and response groups ${\{o_i\}}_{i=1}^G$ generated by the behavior policy $\pi_{\theta_{\mathrm{old}}}$. For each response $o_i$, the objective averages over token positions $t$ and applies PPO-style clipping to the importance ratio
\[
\frac{\pi_\theta(o_{i,t} \mid q, o_{i,<t})}
     {\pi_{\theta_{\mathrm{old}}}(o_{i,t} \mid q, o_{i,<t})}
\]
within the interval $[1-\varepsilon, 1+\varepsilon]$, scaled by $\hat{A}_{i,t}$. This retains PPO's trust-region-like behavior while replacing a parametric critic with group-normalized rewards.

GRPO further adds an explicit KL regularization term between the current policy $\pi_\theta$ and a reference policy $\pi_{\mathrm{ref}}$, weighted by $\gamma$ in Eq.~\eqref{eq:grpo}. The corresponding divergence is estimated at the token level using the non-negative, unbiased Monte Carlo estimator of \citet{schulman2020_klapprox}:
\[
\widehat{D}_{\mathrm{KL}}\!\left(\pi_\theta \,\|\, \pi_{\text{ref}}\right)
= r_{i,t} - \log r_{i,t} - 1
\]
where
\[
r_{i,t}
= \frac{\pi_{\mathrm{ref}}\bigl(o_{i,t} \mid q, o_{i,<t}\bigr)}
        {\pi_{\theta}\bigl(o_{i,t} \mid q, o_{i,<t}\bigr)}
\]

\begin{figure*}[t]
\centering
\begin{tcolorbox}[grpoeqbox, width=\textwidth]
\begin{equation}
\label{eq:grpo}
\resizebox{0.94\textwidth}{!}{$
\begin{aligned}
\mathcal{J}_{\mathrm{GRPO}}(\theta)
&=
\mathbb{E}\bigl[q \sim P(Q),\, \{o_i\}_{i=1}^G \sim \pi_{\theta_{\mathrm{old}}}(O \mid q)\bigr]
\\[0.6em]
&\quad
\frac{1}{G} \sum_{i=1}^{G} \frac{1}{|o_i|} \sum_{t=1}^{|o_i|}
\Bigg\{
\min\Bigg[
\frac{\pi_\theta(o_{i,t} \mid q, o_{i,<t})}
     {\pi_{\theta_{\mathrm{old}}}(o_{i,t} \mid q, o_{i,<t})}
\hat{A}_{i,t},\;
\operatorname{clip}\Bigg(
\frac{\pi_\theta(o_{i,t} \mid q, o_{i,<t})}
     {\pi_{\theta_{\mathrm{old}}}(o_{i,t} \mid q, o_{i,<t})},
1-\varepsilon,\,1+\varepsilon
\Bigg)\hat{A}_{i,t}
\Bigg]
- \gamma\, \mathds{D}_{\mathrm{KL}}\bigl[\pi_\theta \,\|\, \pi_{\mathrm{ref}}\bigr]
\Bigg\}
\end{aligned}
$}
\end{equation}
\end{tcolorbox}
\vspace{-0.6cm}
\end{figure*}

\begin{figure*}[t]
\centering
\begin{tcolorbox}[grpoeqbox, width=\textwidth]
\begin{equation}
\label{eq:dpo}
\resizebox{0.65\textwidth}{!}{$
\mathcal{J}_{\mathrm{DPO}}(\pi_\theta; \pi_{\mathrm{ref}})
=
\mathbb{E}_{(q, y_w, y_l) \sim \mathcal{D}}
\left[
    \log \sigma\!\left(
        \beta \log \frac{\pi_\theta(y_p \mid q)}{\pi_{\mathrm{ref}}(y_p \mid q)}
        -
        \beta \log \frac{\pi_\theta(y_r \mid q)}{\pi_{\mathrm{ref}}(y_r \mid q)}
    \right)
\right]
$}
\end{equation}
\end{tcolorbox}
\vspace{-0.5cm}
\end{figure*}

\noindent\textbf{DPO} trains a target policy $\pi_\theta$ to match human preference judgments while remaining close to a fixed reference policy $\pi_{\mathrm{ref}}$. Training proceeds in two stages. For each query $q$, we first sample a pair of candidate completions $y_1, y_2 \sim \pi_{\mathrm{ref}}(\cdot \mid q)$ from the reference policy and obtain a human label indicating which completion is preferred. This yields an offline preference dataset
\[
\mathcal{D}
=
\bigl\{ \bigl(q^{(i)}, y^{(i)}_{\mathrm{p}}, y^{(i)}_{\mathrm{r}}\bigr) \bigr\}_{i=1}^N
\]
where $y^{(i)}_{\mathrm{p}}$ and $y^{(i)}_{\mathrm{r}}$ denote the preferred and rejected responses to the $i$-th query, respectively. Given $\mathcal{D}$, the frozen reference policy $\pi_{\mathrm{ref}}$, and a scale parameter $\beta$, DPO optimizes $\pi_\theta$ by maximizing the objective $\mathcal{J}_{\mathrm{DPO}}$ in Eq.~(\ref{eq:dpo}). For each triple $(q, y_{\mathrm{p}}, y_{\mathrm{r}})$, the argument of the logistic function $\sigma(\cdot)$ is the $\beta$-scaled difference between the log-likelihood ratios of the preferred and rejected completions, both measured relative to $\pi_{\mathrm{ref}}$. Maximizing $\mathcal{J}_{\mathrm{DPO}}$ therefore upweights preferred responses and downweights rejected ones, while $\beta$ acts as a temperature-like hyperparameter that controls how far $\pi_\theta$ is allowed to move away from $\pi_{\mathrm{ref}}$ to fit the observed preferences, balancing preference alignment against proximity to the reference model.

\subsection{AMIR-GRPO}
AMIR-GRPO augments GRPO by leveraging relative supervision within each rollout group. Rather than treating rewards as isolated scalars, we interpret their within-group ranking as implicit pairwise feedback and use it to refine the policy update. Figure~\ref{fig:pipeline} provides a schematic of AMIR-GRPO: group-wise rewards induce an implicit preference set, which is enforced by a DPO-style contrastive term and optimized jointly with GRPO.

Given a query $q \sim P(Q)$, GRPO samples a group of $G$ completions $\{o_i\}_{i=1}^G$ with scalar rewards $\{r_i\}_{i=1}^G$. A naive construction would include all ordered pairs, yielding $O(G^2)$ preferences per group. To discard near ties and retain only reliable comparisons, we define the implicit preference set
\begin{equation*}
    \mathcal{S}(x) 
    = \left\{ (i,j) \,\middle|\,
        r_i > r_j, \ 
        r_i - r_j > \delta_r
      \right\}
    \label{eq:pair-set}
\end{equation*}
where $\delta_r > 0$ is a reward margin. Each completion may participate in multiple comparisons, so the information within each group of rollouts is exploited more thoroughly. In particular, low-reward completions appear repeatedly as the negative side of the comparison, amplifying their influence beyond what a single scalar advantage can express and leading to stronger suppression of consistently poor reasoning paths.

For each $(i,j) \in \mathcal{S}(q)$ we treat $o_i$ as preferred and $o_j$ as rejected, and introduce the DPO-style logit
\begin{equation*}
\resizebox{0.45\textwidth}{!}{$
    z_{i,j}(\theta)
    = \beta_\text{DPO}
      \Big[
        \big(\ell_{\theta}(q, o_i) - \ell_{\text{ref}}(q, o_i)\big)
        - \big(\ell_{\theta}(q, o_j) - \ell_{\text{ref}}(q, o_j)\big)
      \Big]
    \label{eq:dpo-logit}
$}
\end{equation*}
where $\ell_{\theta}(q,o)$ and $\ell_{\text{ref}}(q,o)$ denote the length-normalized log-probabilities of completion $o$ under the current and reference policies, and $\beta_\text{DPO}$ is a temperature. The associated preference regularizer is
\begin{equation*}
    \mathcal{J}_{\text{pref}}(\theta)
    = \mathbb{E}_{(q,i,j) \sim \mathcal{S}}
      \big[
        \log \sigma\big(z_{i,j}(\theta)\big)
      \big]
\end{equation*}
where $\mathcal{S} = {(q,i,j) \mid (i,j) \in \mathcal{S}(q)}$ is the union of all per-query preference sets. In practice, for a minibatch of queries we approximate this expectation by a simple average over all mined pairs in the batch
\begin{equation*}
    \mathcal{J}_{\text{pref}}(\theta)
    \approx
    \frac{1}{\sum_{q} |\mathcal{S}(q)|}
    \sum_{q}
    \sum_{(i,j) \in \mathcal{S}(q)}
      \log \sigma\big(z_{i,j}(\theta)\big)
\end{equation*}
so that maximizing $\mathcal{J}_{\text{pref}}$ encourages the policy to assign higher log-probability to preferred completions than to rejected ones, with an effective margin controlled by $\beta_\text{DPO}$.

The overall AMIR-GRPO objective augments the standard GRPO loss with this preference term:
\begin{equation*}
    \mathcal{J}_{\text{AMIR-GRPO}}(\theta)
    = \mathcal{J}_{\text{GRPO}}(\theta)
      + \lambda_{\text{reg}} \,\mathcal{J}_{\text{pref}}(\theta)
    \label{eq:amir-grpo}
\end{equation*}
where $\lambda_{\text{reg}} > 0$ scales the contrastive regularizer. A fixed $\lambda_{\text{reg}}$ may either under-utilize the preference signal or overwhelm the base GRPO objective, so we adapt it online to maintain a stable ratio between their contributions. Specifically, after computing both the GRPO loss and the contrastive preference loss on each mini-batch, we track the regularizer's relative contribution to the total objective and update $\lambda_{\text{reg}}$ multiplicatively to keep this ratio within a target range. If the contrastive term becomes too weak relative to the policy loss, we increase $\lambda_{\text{reg}}$ to strengthen the influence of preferences; if it becomes too strong, we reduce $\lambda_{\text{reg}}$ to avoid driving optimization toward pure preference fitting.


\section{Experimental Setup}

\subsection{Datasets}
Our models are trained on the training split of GSM8K \citep{cobbe2021training}, a benchmark of grade-school word problems designed to probe multi-step mathematical reasoning.

For evaluation, we employ a broad suite of math-centric benchmarks: the GSM8K test split \citep{cobbe2021training}, the 2025 American Invitational Mathematics Examination (AIME 2025) \citep{aime25}, OlympiadBench \citep{he-etal-2024-olympiadbench}, AMC23 \citep{AMC23}, Minerva \citep{lewkowycz2022solving}, AQUA-RAT \citep{ling-etal-2017-program}, and LiveMathBench \citep{liu-etal-2025-llms-capable}. Collectively, these datasets span diverse difficulty levels and problem formats, allowing us to assess generalization beyond GSM8K and robustness under more challenging mathematical settings.

\subsection{Training Protocol}
We evaluate our approach by fine-tuning Qwen2.5-Instruct (3B and 7B) \citep{qwen2025qwen25technicalreport} and Gemma3-it (4B) \citep{team2025gemma} under a unified training configuration, comparing the standard GRPO objective with our AMIR-GRPO variant. We apply LoRA with rank 16 and scaling factor $\alpha = 16$, a maximum sequence length of 2048 tokens, a learning rate of $5\times10^{-6}$, and the AdamW optimizer, and use a completion group size of 8.


The reward functions provide the learning signal for our policy optimization. We employ three complementary components: (i) a \emph{correctness reward} that verifies whether the final prediction matches the ground-truth answer, (ii) a \emph{format reward}, which encourages adherence to the prescribed structured output format; and (iii) a \emph{calibration reward}, which incentivizes well-calibrated confidence estimates. A trustworthy prediction system should produce confidence scores that are calibrated, i.e., the reported confidence should coincide with the true probability that the prediction is correct. To operationalize this, we use a \textit{Verbalized Confidence} protocol \citep{tian-etal-2023-just}: for each question--response pair, the model is prompted to output both an answer and an explicit confidence value $q \in [0,1]$. We then calibrate this score by minimizing a strictly proper scoring rule $S$ with respect to the binary correctness indicator $a\in\{0,1\}$; in expectation, $S$ is uniquely minimized when the reported confidence matches the true success probability $p(a=1)$:
\[
\mathbb{E}_{a \sim p(a)}\bigl[S(p(a), a)\bigr]
\;\leq\;
\mathbb{E}_{a \sim p(a)}\bigl[S(q, a)\bigr]
\]

In our implementation, we instantiate $S$ as the Brier score \citep{glenn1950verification}:
\[
S(q,a) = (a - q)^2
\]
where $a \in \{0,1\}$ indicates correctness. The resulting \emph{calibration reward} is defined as its complement, $1 - S(q,a)$. The overall reward $R$ is then given by a weighted sum of the three components:
\[
R \;=\; w_{\text{corr}}\, r_{\text{corr}} \;+\; w_{\text{fmt}}\, r_{\text{fmt}} \;+\; w_{\text{calib}}\, r_{\text{calib}} \footnote{In our experiments $(w_{\text{corr}}, w_{\text{fmt}}, w_{\text{calib}})=(2.0,\,0.9,\,1.0)$}
\]

\subsection{Evaluation Metric}
Conventional evaluation protocols aggregate scores from greedy decoding or nucleus sampling \citep{Holtzman2020The}, thereby reflecting only average-case behavior. This can underestimate a model's true capability: a system may fail on a difficult instance under a small sampling budget even though additional samples would yield a correct solution. To address this, we adopt the Pass@$k$ metric.

A naive computation of Pass@$k$ from exactly $k$ samples per problem suffers from high variance. We therefore employ the unbiased estimator introduced by \citet{chen2021evaluatinglargelanguagemodels}, which measures, for a given question $q$, the probability that at least one correct solution appears among $k$ sampled generations:
\begin{equation*}
\text{Pass@}k = \mathbb{E}_{q}\Bigg[ 1 - \frac{\binom{n - c_q}{k}}{\binom{n}{k}} \Bigg]
\end{equation*}
where $n$ is the total number of generations produced for question $q$, and $c_q$ is the number of correct generations among them. Consequently, Pass@$k$ corresponds to the expected fraction of questions for which at least one of the $k$ sampled generations is correct.

Unless stated otherwise, we fix $n = 8$ generations per question and report Pass@$k$ for $k \in \{1, 2, 4\}$.\footnote{Generations are sampled with temperature $0.6$ and top-$p$ sampling with $p = 1.0$.}

\section{Results and Analysis}
Table~\ref{tab:amir_grpo_main} summarizes the main results. Across all evaluated values of $k$, AMIR-GRPO consistently improves out-of-distribution generalization while maintaining GRPO-comparable performance on the in-distribution GSM8K benchmark, yielding a robust balance between accuracy at small $k$ and exploration at large $k$.

These findings support our core claim: implicit-preference regularization provides richer supervision at no additional data cost, increasing the value extracted from existing datasets. Appendix~\ref{app:training-dynamics} details the training dynamics, while Appendix~\ref{app:prompt-templates} provides the full prompt template.

\begin{table}[t]
  \centering
  \setlength{\tabcolsep}{6pt}
  \renewcommand{\arraystretch}{1.2}

  \resizebox{\linewidth}{!}{%
  \begin{tabular}{lcccc}
    \toprule
    \textbf{Dataset} & \boldmath$k$ &
    \textbf{Qwen2.5-3B-Instruct} &
    \textbf{Qwen2.5-7B-Instruct} &
    \textbf{Gemma-3-4B-it} \\
    \midrule

    \multirow{3}{*}{GSM8K}
      & 1 & \reduction{83.6}{83.4}{-0.2} & \reduction{90.4}{90.2}{-0.2} & \entry{86.2}{86.4}{+0.2} \\
      & 2 & \reduction{90.2}{89.9}{-0.3} & \reduction{94.6}{94.3}{-0.3} & \entry{90.3}{90.7}{+0.4} \\
      & 4 & \reduction{93.6}{93.4}{-0.2} & \reduction{96.4}{96.2}{-0.2} & \entry{92.8}{93.2}{+0.4} \\
    \midrule

    \multirow{3}{*}{AIME25}
      & 1 & \entry{0.8}{2.1}{+1.3} & \entry{4.2}{5.8}{+1.6} & \entry{3.7}{5.8}{+2.1} \\
      & 2 & \entry{1.7}{4.2}{+2.5} & \entry{7.5}{9.6}{+2.1} & \entry{4.9}{8.7}{+3.8} \\
      & 4 & \entry{3.3}{8.3}{+5.0} & \entry{12.1}{13.8}{+1.7} & \entry{5.9}{12.4}{+6.5} \\
    \midrule

    \multirow{3}{*}{OlympiadBench}
      & 1 & \entry{26.4}{27.0}{+0.6} & \entry{36.2}{36.4}{+0.2} & \entry{22.2}{22.7}{+0.5} \\
      & 2 & \entry{33.8}{34.5}{+0.7} & \entry{43.6}{43.7}{+0.1} & \entry{27.3}{28.1}{+0.8} \\
      & 4 & \entry{40.9}{42.4}{+1.5} & \entry{50.5}{50.6}{+0.1} & \entry{31.6}{33.0}{+1.4} \\
    \midrule

    \multirow{3}{*}{AMC23}
      & 1 & \entry{30.9}{32.8}{+1.9} & \entry{45.0}{47.8}{+2.8} & \entry{29.4}{35.0}{+5.6} \\
      & 2 & \entry{40.5}{43.2}{+2.7} & \entry{55.4}{60.9}{+5.5} & \entry{35.4}{41.8}{+6.4} \\
      & 4 & \entry{52.3}{53.5}{+1.2} & \entry{63.7}{73.1}{+9.4} & \entry{39.2}{48.2}{+9.0} \\
    \midrule

    \multirow{3}{*}{MinervaMath}
      & 1 & \entry{25.2}{26.3}{+1.1} & \entry{35.2}{35.9}{+0.7} & \entry{20.9}{21.9}{+1.0} \\
      & 2 & \entry{32.6}{33.7}{+1.1} & \entry{41.6}{42.8}{+1.2} & \entry{26.9}{28.3}{+1.4} \\
      & 4 & \entry{38.9}{40.3}{+1.4} & \entry{46.5}{48.4}{+1.9} & \entry{32.3}{33.7}{+1.4} \\
    \midrule

    \multirow{3}{*}{AQUA-RAT}
      & 1 & \entry{52.6}{54.0}{+1.4} & \entry{59.0}{60.0}{+1.0} & \entry{50.8}{51.9}{+1.1} \\
      & 2 & \entry{61.0}{62.0}{+1.0} & \entry{66.0}{66.5}{+0.5} & \entry{56.4}{57.5}{+1.1} \\
      & 4 & \entry{66.7}{67.3}{+0.6} & \entry{70.4}{70.8}{+0.4} & \entry{61.4}{62.3}{+0.9} \\
    \midrule

    \multirow{3}{*}{LiveMathBench}
      & 1 & \entry{19.0}{23.1}{+4.1} & \entry{27.4}{30.2}{+2.8} & \entry{13.9}{15.8}{+1.9} \\
      & 2 & \entry{24.1}{31.6}{+7.5} & \entry{34.5}{38.4}{+3.9} & \entry{17.4}{19.0}{+1.6} \\
      & 4 & \entry{27.3}{39.2}{+11.9} & \entry{41.7}{45.1}{+3.4} & \entry{20.9}{22.1}{+1.2} \\
    \bottomrule
  \end{tabular}%
  }

  \caption{Comparative performance on math benchmarks. Numbers are reported in \%; Cells are formatted as \textcolor{basecolor}{GRPO} $\to$ \textbf{AMIR-GRPO} \textcolor{gaincolor}{(+gain)}.}
  \label{tab:amir_grpo_main}
  \vspace{-0.5cm}
\end{table}

\subsection{\texorpdfstring{$\beta_\text{DPO}$}{beta	extsubscript{DPO}} Sensitivity}

\begin{figure*}[t]
    \centering
    \includegraphics[width=0.9\textwidth]{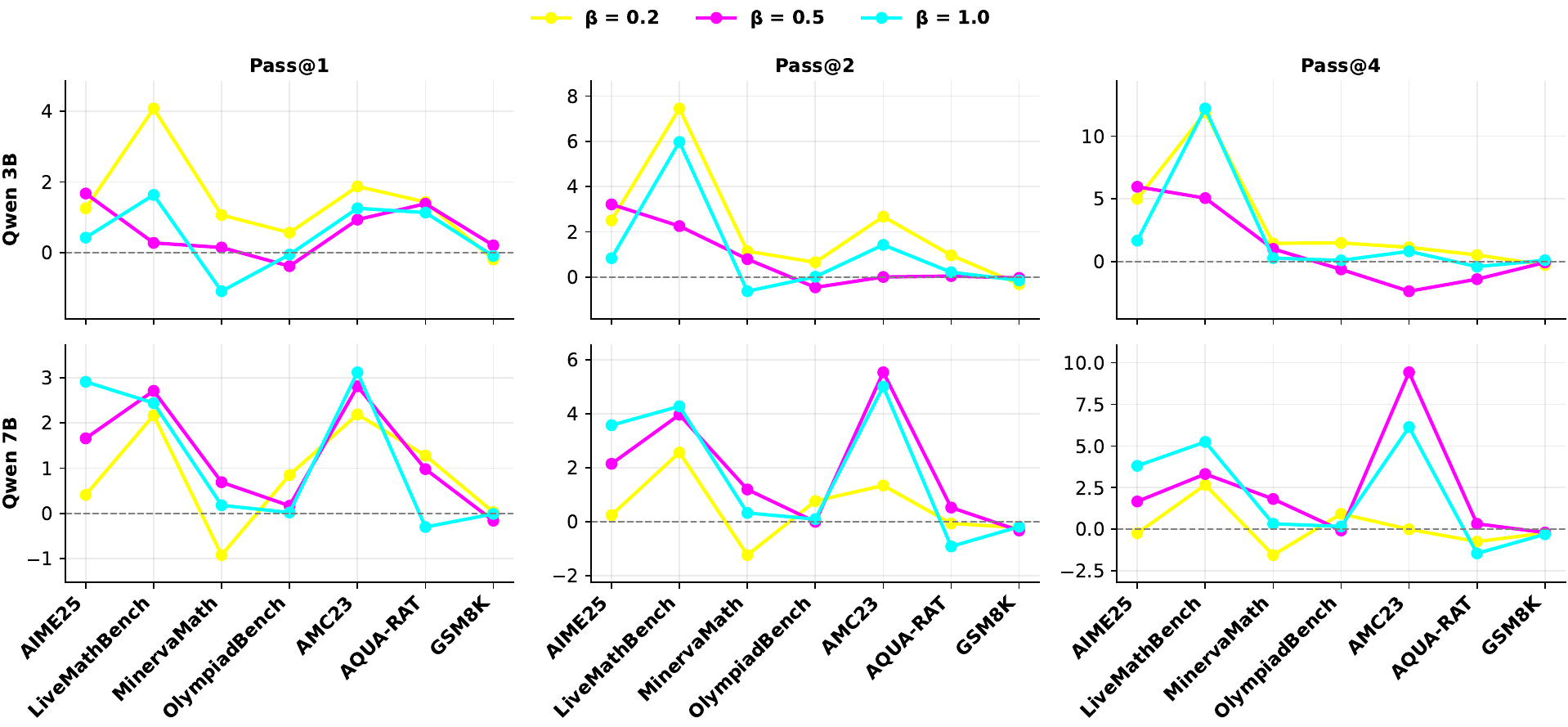}
    \caption{Impact of $\beta_\text{DPO}$ across datasets. Values indicate $\Delta\text{Pass@}k$ relative to the GRPO baseline.}
    \vspace{-0.3cm}
    \label{fig:beta-effect}
\end{figure*}

In the AMIR-GRPO objective, $\beta_\text{DPO}$ acts as a temperature for the DPO-style contrastive preference regularizer $\mathcal{J}_{\text{pref}}(\theta)$, controlling the effective margin by which the policy distinguishes high- from low-reward completions. Adding $\mathcal{J}_{\text{pref}}(\theta)$ consistently improves performance, but the magnitude of the gain is sensitive to $\beta_\text{DPO}$ and is jointly governed by the base model’s capacity, initial competence and the dataset’s difficulty.

For small, initially weak models (e.g., Qwen-3B), smaller $\beta_\text{DPO}$ is preferable: limited capacity makes them prone to high-variance preference gradients, so a gentle contrastive signal stabilizes training and prevents $\mathcal{J}_{\text{pref}}$ from dominating the core $\mathcal{J}_{\text{GRPO}}$ term. Larger models with higher capacity but similarly low initial competence (e.g., Qwen-7B) tolerate stronger alignment pressure and typically peak at moderate $\beta_\text{DPO}$
, where the induced margin is decisive without becoming overconfident. For already strong base models (e.g., Gemma-4B), higher $\beta_\text{DPO}$ tends to be more beneficial.

Even within a fixed architecture, the optimal value remains dataset-dependent. For our 7B policy, a moderate $\beta_\text{DPO}$ performs best on average, yet the most demanding benchmarks (e.g., AIME25 and LiveMathBench) favor larger settings—suggesting that a wider preference margin helps capitalize on subtle reward differences in complex regimes. Overall, these results underscore the need to tune $\beta_\text{DPO}$ jointly to the model and data to fully unlock AMIR-GRPO’s alignment benefits; the corresponding trends are summarized in Figure~\ref{fig:beta-effect}.

\subsection{Are RL Reasoning Paths Truly Novel?}

\begin{figure*}[t]
    \centering
    \includegraphics[width=0.9\textwidth]{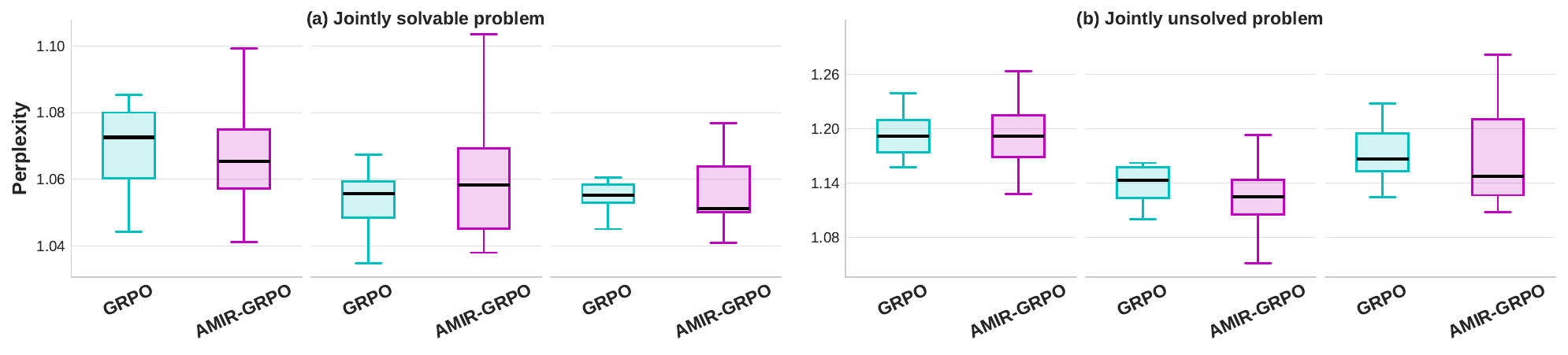}
    \vspace{-0.3cm}
    \caption{Comparison of reasoning-trajectory perplexity on an AMC23 question.}
    \label{fig:perplexity_analysis}
    \vspace{-0.5cm}
\end{figure*}

A recurring question in RL-based reasoning is whether performance gains stem from truly novel reasoning trajectories or from reweighting trajectories that were already plausible under the base model. \citet{yue2025does} study this by analyzing conditional perplexity of generated responses and conclude that reinforcement learning largely sharpens probability mass around pre-existing modes rather than discovering fundamentally new reasoning patterns. In the same vein, we examine how AMIR-GRPO reshapes the distribution of reasoning chains relative to GRPO, using perplexity as a diagnostic of how well a sampled rollout is supported by the very policy that generated it.

Formally, for a model $m$, a problem $x$, and a response $Y=(y_1,\dots,y_T)$ of length $T$ tokens, we define conditional perplexity as the exponentiated average negative log-likelihood of $Y$ under $m$ \citep{yue2025does}:
\begin{equation*}
\resizebox{0.99\columnwidth}{!}{$
\displaystyle
\mathrm{PPL}_m(Y \mid x)
= \exp\Biggl(
    -\frac{1}{T} \sum_{t=1}^{T} 
    \log P_m\bigl(y_t \mid x, y_1,\dots,y_{t-1}\bigr)
\Biggr)
$}
\end{equation*}
Lower values of $\mathrm{PPL}_m(Y \mid x)$ signify that $Y$ lies deeper in high-density regions of $m$’s conditional distribution, i.e., a more self-consistent and strongly supported trajectory.

Using Gemma-4B, we analyze six AMC23 problems randomly selected from two pre-filtered strata: jointly solvable instances, where both tuned models (GRPO and AMIR-GRPO) consistently produce correct chains of thought across samples, and jointly unsolved instances, where all sampled trajectories are incorrect for both models. For each problem and each model, we sample $16$ rollouts and compute $\mathrm{PPL}_m(Y \mid x)$ under the same model $m$ that generated $Y$. Figure~\ref{fig:perplexity_analysis} demonstrates the resulting distributions. As expected, jointly unsolved problems exhibit higher perplexity than jointly solvable ones, reflecting the greater challenge of generating chains that are internally supported in this regime. 

Across both panels, GRPO typically produces tighter boxes with shorter whiskers, suggesting that it samples from a comparatively narrow band of rollouts that are similarly likely under its own policy—consistent with a more sharply mode-concentrated trajectory distribution. In contrast, AMIR-GRPO generally matches or slightly improves the median perplexity relative to GRPO, while exhibiting a wider interquartile range and longer whiskers. In other words, AMIR-GRPO often includes rollouts that are as self-likely as GRPO’s typical samples (low perplexity), yet it also permits a nontrivial fraction of trajectories that the policy itself assigns lower probability (higher perplexity).

Because perplexity is evaluated under the generating model, variation in $\mathrm{PPL}_m$ directly reflects how peaked the policy is over the trajectories it actually produces. GRPO’s tight clustering is consistent with strong RL-induced sharpening: once a small set of high-reward trajectories is identified, probability mass concentrates around a narrow mode, making most sampled chains uniformly high-likelihood under the GRPO policy. AMIR-GRPO retains some of this sharpening—its medians are frequently comparable or lower—but its regularization term curbs excessive collapse. The resulting distribution is effectively tempered: AMIR-GRPO can prioritize effective reasoning patterns (low median perplexity) while sustaining a thicker tail and broader interquartile range (greater trajectory diversity). This behavior is the expected signature of a regularized objective: it shifts mass toward stronger chains without extinguishing alternative, lower-probability reasoning paths that support exploration and coverage.

\subsection{Solvable Problem Space}
To further investigate whether RL training truly enlarges the set of solvable tasks or primarily reweights trajectories already available to the Base model (Qwen-3B). On AMC23, we define a question as solvable if at least one of 16 sampled rollouts yields the correct answer, and compare solvable sets across the Base model and its GRPO- and AMIR-GRPO-trained variants. As summarized in Table~\ref{tab:solvable_problem_coverage}, 70.6\% of solvable questions lie in the intersection of all three systems, revealing a substantial shared core of problems that remain accessible regardless of the downstream objective. This overlap suggests that a large portion of the observed RL gains stem from reshaping probability mass over trajectories that were already realizable under the Base model, rather than from discovering entirely new modes of reasoning.

The remaining problems, however, highlight a clear contrast between the two RL objectives. Only 2.9\% of solvable questions are handled by the Base model and the standard GRPO variant but not by AMIR-GRPO, and there are no questions uniquely solved by GRPO when both the Base and AMIR-GRPO fail. By contrast, AMIR-GRPO delivers a nontrivial expansion of coverage: 11.8\% of solvable problems are addressed jointly by the Base and AMIR-GRPO but missed by GRPO, and an additional 8.8\% are solved exclusively by AMIR-GRPO when both the Base and GRPO fail. Very few problems (2.9\%) are solvable only by the Base model, indicating that neither RL configuration catastrophically erodes the Base model’s capabilities.


\begin{table}[h!]
  \centering
  \setlength{\tabcolsep}{5pt}
  \renewcommand{\arraystretch}{0.75}%
  \resizebox{0.8\columnwidth}{!}{%
    \begin{tabular}{ccc|c}
      \toprule
     {\small \textbf{Base}} & {\small \textbf{GRPO}} & {\small \textbf{AMIR-GRPO}} & {\small \textbf{Coverage}} \\
      \midrule
      \cmark & \cmark & \cmark & {\small70.6\%} \\
      \cmark & \cmark & \xmark & {\small2.9\%} \\
      \cmark & \xmark & \cmark & {\small11.8\%} \\
      \cmark & \xmark & \xmark & {\small2.9\%} \\
      \midrule
      \xmark & \cmark & \cmark & {\small2.9\%} \\
      \xmark & \cmark & \xmark & {\small0.0\%} \\
      \xmark & \xmark & \cmark & {\small8.8\%} \\
      \xmark & \xmark & \xmark & {\small0.0\%} \\
      \bottomrule
    \end{tabular}
  }
  \caption{Coverage of solvable AMC23 problems.}
  \vspace{-0.4cm}
  \label{tab:solvable_problem_coverage}
\end{table}

\subsection{Sharper Margins, Stronger Reasoning}
On Gemma-4B evaluated on LiveMathBench, we analyze a post-hoc preference margin defined as the per-problem difference between the mean length-normalized log-probability assigned to correct versus incorrect completions. This scalar margin is positive when the model, on average, assigns higher per-token likelihood to correct reasoning trajectories than to incorrect ones. Figure~\ref{fig:margin_diff} shows that both GRPO and AMIR-GRPO achieve positive margins, but AMIR-GRPO yields a noticeably larger average margin ($\times$ 2.7) compared to GRPO, indicating a sharper and more confident separation between correct and incorrect solutions in probability space.

\begin{figure}[h]
    \centering
    \includegraphics[width=0.7\columnwidth]{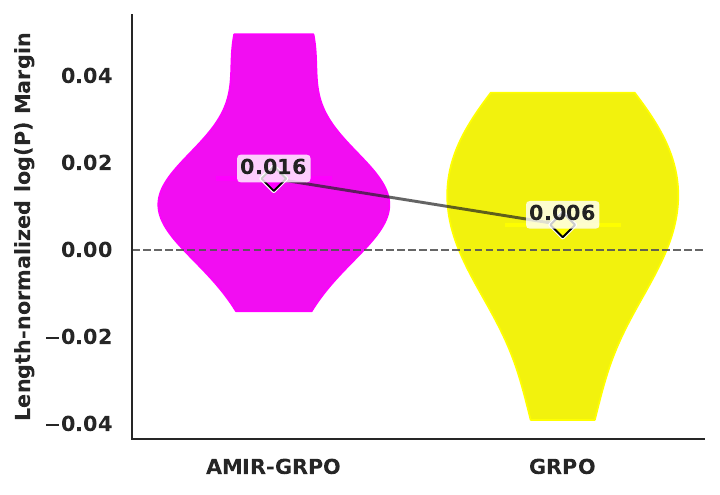}
    \vspace{-0.1cm}
    \caption{Preference margin separation.}
    \label{fig:margin_diff}
    \vspace{-0.5cm}
\end{figure}

\subsection{Dissection of Reasoning Failures}
We go beyond accuracy to characterize how models fail, conducting a fine-grained error analysis on AMC23. For each problem, we sample 16 rollouts from Qwen-7B and convert every solution into a stepwise form by segmenting it into numbered reasoning steps. For each incorrect rollout, we identify the first erroneous step\footnote{the earliest point at which the reasoning irreversibly departs from a valid solution} and assign an error label using our task-specific taxonomy (Appendix~\ref{app:error_taxonomy}). Labels are initially proposed by GPT-5 \citep{openai_gpt5_docs_2025} and then manually audited to ensure consistency and avoid systematic misclassification.

\begin{figure}[h]
    \centering
    \includegraphics[width=0.95\columnwidth]{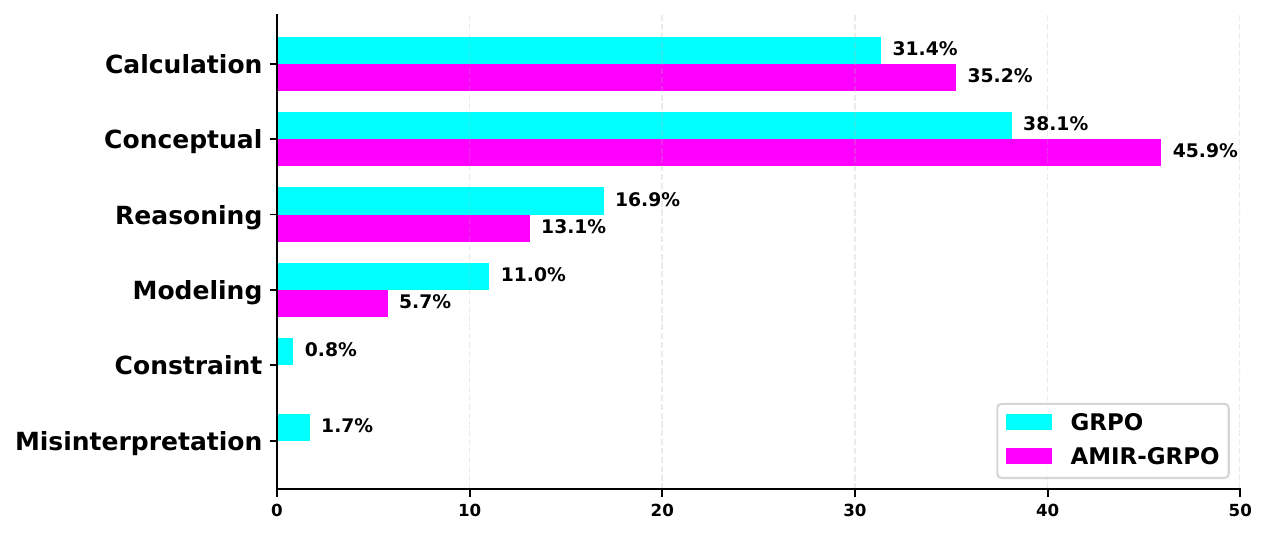}
    \vspace{-0.1cm}
    \caption{Error distribution.}
    \label{fig:error_types}
    \vspace{-0.3cm}
\end{figure}

Figure~\ref{fig:error_types} summarizes the resulting error-type distribution, revealing the most prevalent failure modes. Complementarily, Figure~\ref{fig:failure_locality_density} examines where breakdowns occur within the reasoning chain. For each incorrect solution, we compute a relative error position\footnote{(step index)/(number of steps)} and estimate a smoothed density over all failures. This tests whether errors tend to arise early (problem interpretation/formulation), mid-chain (intermediate inference), or late (final execution).

\begin{figure}[h]
    \centering
    \includegraphics[width=0.95\columnwidth]{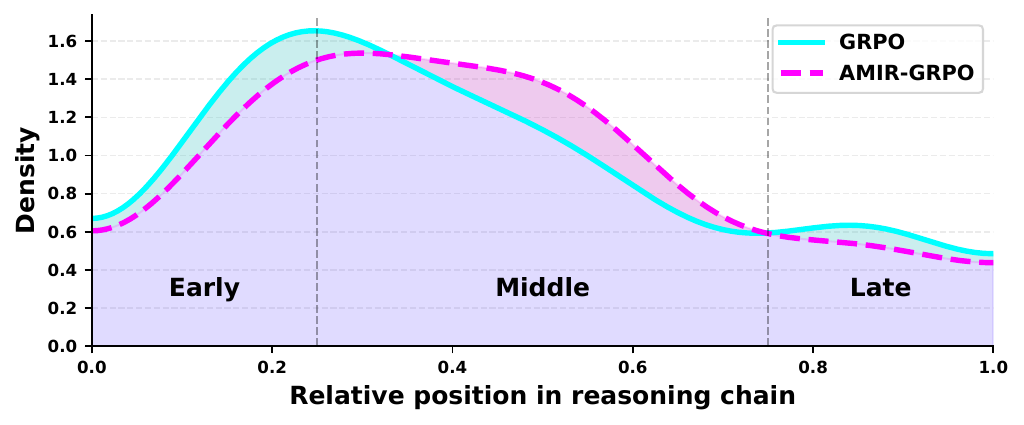}
    \vspace{-0.2cm}
    \caption{Failure locality density.}
    \label{fig:failure_locality_density}
    \vspace{-0.55cm}
\end{figure}

Comparing GRPO to AMIR-GRPO, Figure~\ref{fig:failure_locality_density} shows a clear shift: AMIR-GRPO fails less frequently at both the start and the end of the chain. This suggests that AMIR-GRPO reduces formulation-related errors (early-stage) and answer realization errors (late-stage), implying that its improvements reflect not only a higher success rate but also a systematic change in where reasoning tends to break when it fails.

\subsection{Cross-Variant Integratability}
Our approach is designed to integrate seamlessly with other GRPO-family variants, including GSPO \citep{zheng2025group}, offering robust compatibility. Notably, it delivers consistent gains across a wide suite of benchmarks, as reported in Table~\ref{tab:gspo_results}.

\begin{table}[h]
  \centering
  \small
  \setlength{\tabcolsep}{5pt}
  \renewcommand{\arraystretch}{1.25} 

  \resizebox{\columnwidth}{!}{%
    \begin{tabular}{l ccc c}
      \toprule
      \multirow{2}{*}{\textbf{Dataset}}
      & \multicolumn{3}{c}{\textbf{Qwen2.5-7B-Instruct}} \\
      \cmidrule{2-4} 
      & Pass@1 & Pass@2 & Pass@4  \\
      \midrule
      
      GSM8K
        & \entry{90.3}{90.5}{+0.2} & \entry{94.3}{94.3}{+0.0} & \entry{95.8}{96.0}{+0.2} \\
        
      AIME25
        & \entry{3.7}{7.1}{+3.4} & \entry{7.1}{10.5}{+3.4} & \entry{12.9}{14.3}{+1.4} \\

        
      AMC23
        & \entry{45.6}{47.5}{+1.9} & \entry{56.5}{59.7}{+3.2} & \entry{66.2}{70.1}{+3.9} \\
        
      MinervaMath
        & \entry{35.0}{35.5}{+0.5} & \entry{41.5}{41.9}{+0.4} & \entry{46.2}{47.5}{+1.3} \\
        
      AQUA-RAT
        & \entry{60.2}{61.0}{+0.8} & \entry{67.0}{67.7}{+0.7} & \entry{71.3}{71.8}{+0.5} \\
        
      LiveMathBench
        & \entry{29.1}{33.2}{+4.1} & \entry{37.6}{41.3}{+3.7} & \entry{47.2}{50.0}{+2.8} \\
      
      \bottomrule
    \end{tabular}%
  }
\vspace{-0.1cm}
\caption{Comparative performance on benchmarks. Cells are formatted as: \textcolor{basecolor}{GSPO} $\to$ \textbf{AMIR-GSPO} \textcolor{gaincolor}{(+gain)}.}
\vspace{-0.5cm}
\label{tab:gspo_results}
\end{table}

\section{Related Work}
RL has become a key driver of improved reasoning in LLMs. While PPO remains the standard choice, its actor–critic design requires a value network roughly the size of the policy, sharply increasing memory and compute \citep{schulman2017proximal}. This overhead has motivated critic-free, group-based alternatives that avoid explicit value estimation by exploiting relative rewards across multiple completions of the same prompt.

GRPO substitutes the explicit critic with a normalized reward baseline, delivering stable updates without an auxiliary value model \citep{shao2024deepseekmath}. Subsequent variants refine how group-relative signals are computed and normalized. Dr.GRPO drops token-level averaging and reward normalization to mitigate length- and difficulty-induced optimization bias \citep{liu2025understanding}. GPG similarly normalizes by the group mean to temper gradient variance with minimal overhead \citep{chu2025gpg}. 

A complementary line of work focuses on where group-relative objectives act along the sequence and how they are adapted to long chain-of-thought (CoT) reasoning. GSPO lifts importance sampling and clipping from the token level to whole sequences, improving stability while retaining group-relative rewards \citep{zheng2025group}. SAPO replaces hard PPO-style clipping with a smooth, temperature-controlled trust region, yielding less brittle and more adaptive updates \citep{gao2025soft}. TreeRPO extends GRPO to tree-structured reasoning by assigning group-relative rewards at intermediate nodes, converting sparse trajectory-level feedback into denser step-wise signals \citep{yang2025treerpo}. DAPO introduces clip-Higher to preserve diversity and prevent entropy collapse, dynamic sampling to improve efficiency and stability, a token-level policy-gradient loss adapted to long sequences, and overlong reward shaping to suppress noise from excessively long outputs \citep{yu2025dapo}. Revisiting value-based designs, VAPO proposes a value-centric framework with length-adaptive generalized advantage estimation, improving advantage accuracy and stability while integrating several prior optimization refinements \citep{yue2025vapo}.


Our method moves beyond GRPO variants that mainly adjust value modeling. We preserve GRPO’s critic-free core and add an implicit preference objective defined directly from group rewards. This yields richer supervision at zero additional annotation cost, and integrates cleanly with GRPO-family extensions (e.g., GSPO).


\section{Conclusion}
This work identifies limitations of GRPO, showing that the standard formulation fails to fully exploit the within-group ordering present in rollout batches. We propose AMIR-GRPO, which augments GRPO with an implicit preference objective induced by groupwise rankings, while leaving the data collection pipeline unchanged. Across a spectrum of math-reasoning benchmarks, from school-level to Olympiad-level tasks, AMIR-GRPO consistently outperforms standard GRPO at a reasonable and affordable computational cost. Our analysis further indicates that the method sharpens decision boundaries and expands the set of solvable instances, rather than simply amplifying existing trajectories. Taken together, these results suggest that making implicit preference structure explicit in group-based RL provides a simple, scalable mechanism for strengthening reasoning supervision.

\section*{Limitations}

Although AMIR-GRPO demonstrates strong empirical performance in our experiments, several limitations remain. First, we assess the method exclusively on text-based benchmarks. Accordingly, our conclusions are confined to textual reasoning; extending AMIR-GRPO to other modalities (e.g., code or vision--language reasoning) remains an important avenue for future work, where reward design, noise profiles, and failure modes may differ from those in purely textual tasks.

Second, whereas much prior work on long-horizon, verifiable reasoning focuses mainly on mathematical benchmarks \citep{nan2025ngrpo, zhu2025surprising}, we adopt a broader view by evaluating a wider range of math reasoning benchmarks than is typical. Even so, our evaluation still reflects a single stylistic regime. Broadening the study to additional reasoning domains---including commonsense \citep{sprague2024musr}, cultural reasoning \citep{yari-koto-2025-unveiling}, and open-ended instruction following \citep{dussolle-etal-2025-ifeval}---would better establish how general AMIR-GRPO’s gains are under qualitatively different training signals and evaluation criteria.

Finally, our current formulation relies on trajectory-level outcome rewards without explicit process rewards or fine-grained credit assignment to intermediate reasoning steps. While this design choice simplifies the training pipeline, it may limit the algorithm’s ability to correct early-step errors or to exploit partially correct reasoning traces. Compared with purely binary or final-answer rewards, incorporating richer intermediate signals to guide the reasoning trajectory could improve exploration efficiency and more precisely steer the model toward promising solution paths.


\bibliography{custom}

\newpage
\appendix

\section{Prompt Templates}
\label{app:prompt-templates}
We provide the prompt templates used in our experiments.

\begin{tcolorbox}[
  enhanced,
  breakable,
  colback=teal!3!white,
  colframe=teal!60!black,
  coltitle=black,
  title={},
  fonttitle=\bfseries\small,
  boxed title style={
    colback=white,
    colframe=teal!60!black,
    boxrule=0.5pt,
  },
  attach boxed title to top left={yshift=-1mm,xshift=0mm},
  arc=1mm,
  boxrule=0.5pt,
  left=3mm,
  right=3mm,
  top=2mm,
  bottom=2mm,
  listing only,
  listing engine=listings,
  listing options={style=promptstyle}
]
A conversation between User and Assistant. The user asks a question, and the Assistant solves it.
The Assistant first thinks about the reasoning process in the mind, provides the user with the final answer, 
then analyzes its confidence about the solution and provides the user with its confidence level. 

The confidence level is a number between 0 and 1 (inclusive) enclosed within \textcolor{teal!50!black}{<confidence> </confidence>} tags. 

The final answer is enclosed between \textcolor{teal!50!black}{<answer> </answer>} tags. 

The analysis about confidence and uncertainty is enclosed within \textcolor{teal!50!black}{<analysis> </analysis>} tags. 

The Assistant should reason about its confidence in the solution and its uncertainty in the solution within these tags. 

The final format that must be followed is:

\textcolor{teal!50!black}{<think>} reasoning process here \textcolor{teal!50!black}{</think>}

\textcolor{teal!50!black}{<answer>} \textbackslash boxed\{final answer here\} \textcolor{teal!50!black}{</answer>}

\textcolor{teal!50!black}{<analysis>} analysis about confidence and uncertainty here \textcolor{teal!50!black}{</analysis>}

\textcolor{teal!50!black}{<confidence>} confidence level here (number between 0 and 1) \textcolor{teal!50!black}{</confidence>}
\end{tcolorbox}

\section{Training Dynamics}
\label{app:training-dynamics}

\begin{figure*}[t]
    \centering
    \begin{subfigure}{\textwidth}
        \centering
        \includegraphics[width=\linewidth]{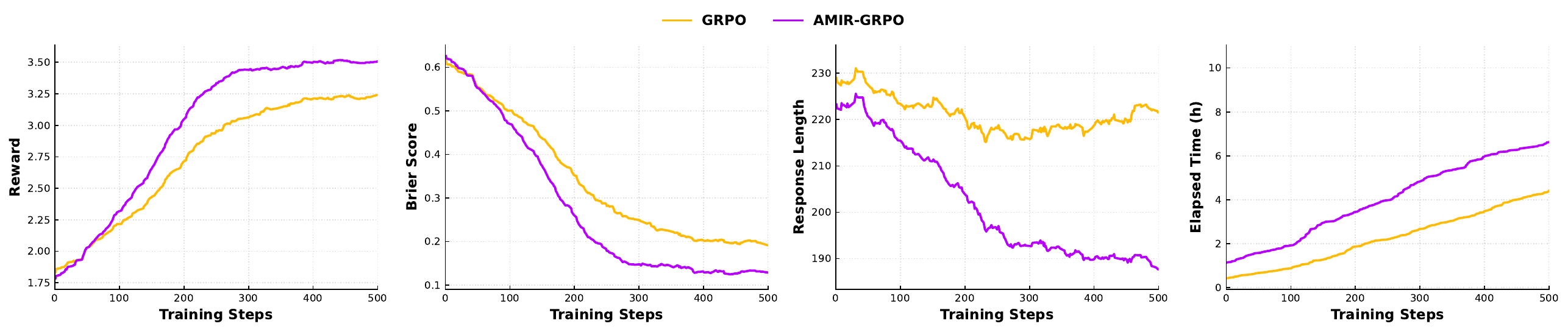}
        \caption{Qwen-3B}
        \label{fig:training_curves_3B}
    \end{subfigure}

    \vspace{0.5em} 

    \begin{subfigure}{\textwidth}
        \centering
        \includegraphics[width=\linewidth]{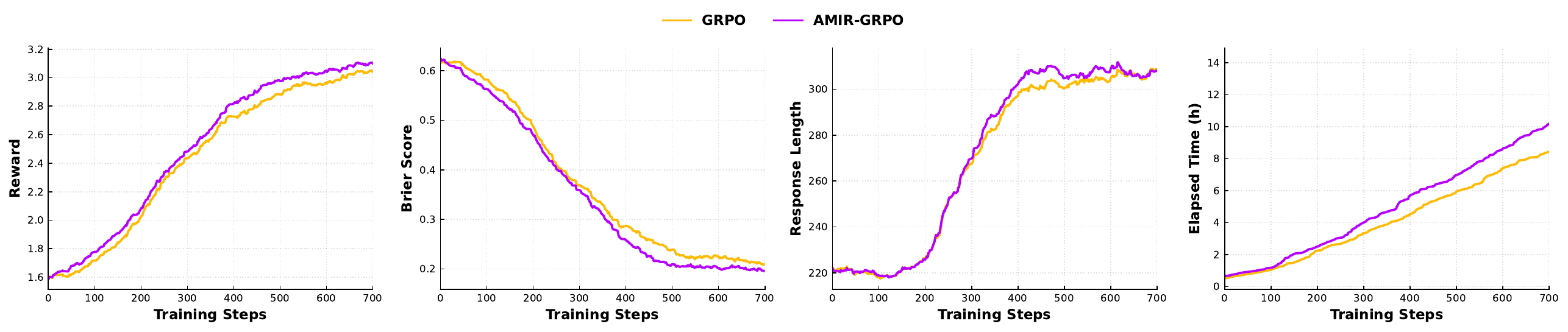}
        \caption{Qwen-7B}
        \label{fig:training_curves_7B}
    \end{subfigure}

    \vspace{0.5em} 

    \begin{subfigure}{\textwidth}
        \centering
        \includegraphics[width=\linewidth]{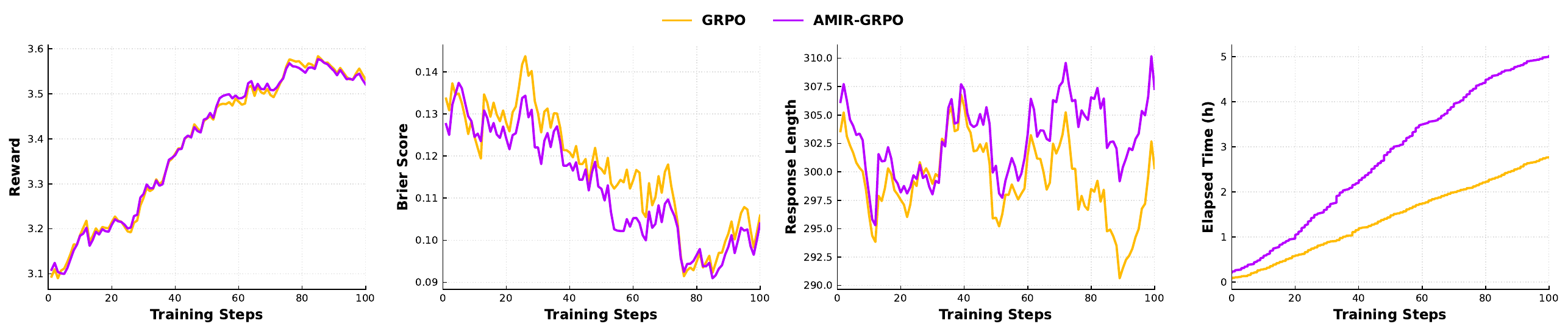}
        \caption{Gemma-4B}
        \label{fig:training_curves_gemma}
    \end{subfigure}

    \caption{Training behavior of GRPO vs. AMIR-GRPO.}
    \label{fig:training_curves}
\end{figure*}

Figure~\ref{fig:training_curves} summarizes the training dynamics of our models. Across all settings, AMIR-GRPO attains consistently lower Brier scores than vanilla GRPO, indicating better-calibrated confidence and a more trustworthy approach.

Its effect on response length is not uniform; it depends on the base model’s capacity and initial competence, and acts as an adaptive regularizer that counterbalances GRPO’s intrinsic bias. For low-capacity, low-competence models (Qwen-3B), it behaves like a pruning operator, sharply shortening outputs by suppressing long, low-reward hallucinations that GRPO’s diluted scalar penalty fails to eliminate. Conversely, for competent models, AMIR-GRPO acts as a policy protector, preventing the detrimental brevity bias from compromising necessary reasoning depth. Specifically, it unshackles latent capacity in high-capacity, initially weaker models (Qwen-7B), allowing them to articulate the full Chain-of-Thought required for task mastery, thereby increasing length. Simultaneously, it stabilizes high-competence policies (Gemma-4B), preserving the robust structure of existing high-quality reasoning against GRPO's pressure to enforce premature token efficiency, which otherwise risks policy collapse. Thus, AMIR-GRPO provides an essential adaptive control: it curtails length only where it signifies policy error, yet preserves or expands it where it is indispensable for solution fidelity.

\section{Response Length Characteristics}
\label{app:length-bias}
To quantify response-level length bias, we sample 16 rollouts per query from Qwen-3B on AIME25 and Qwen-7B on AMC23 after fine-tuning with GRPO or AMIR-GRPO. For AIME25, incorrect Qwen-3B trajectories average 747.7 tokens with GRPO versus 738.9 tokens with AMIR-GRPO. On AMC23, the corresponding averages for Qwen-7B are 781.4 and 759.4 tokens. These shorter low-reward trajectories under AMIR-GRPO support our claim that it mitigates response-level length bias by amplifying negative gradients on long, low-reward trajectories, thereby increasing their effective penalty.

\section{Memory Footprint Analysis}
\label{app:memory-usage}
Figure~\ref{fig:gpu_memory} indicates that AMIR-GRPO incurs only a slight GPU memory overhead relative to standard GRPO. Peak consumption increases by 3.0~GiB for Qwen-7B, 3.0~GiB for Gemma-4B, and 4.9~GiB for Qwen-3B, corresponding to average temporal overheads of just 11.4\%, 9.1\%, and 37.5\%, respectively. Consequently, the memory footprint of AMIR-GRPO remains essentially comparable to that of GRPO.

\begin{figure}[H]
    \centering
    \includegraphics[width=0.95\columnwidth]{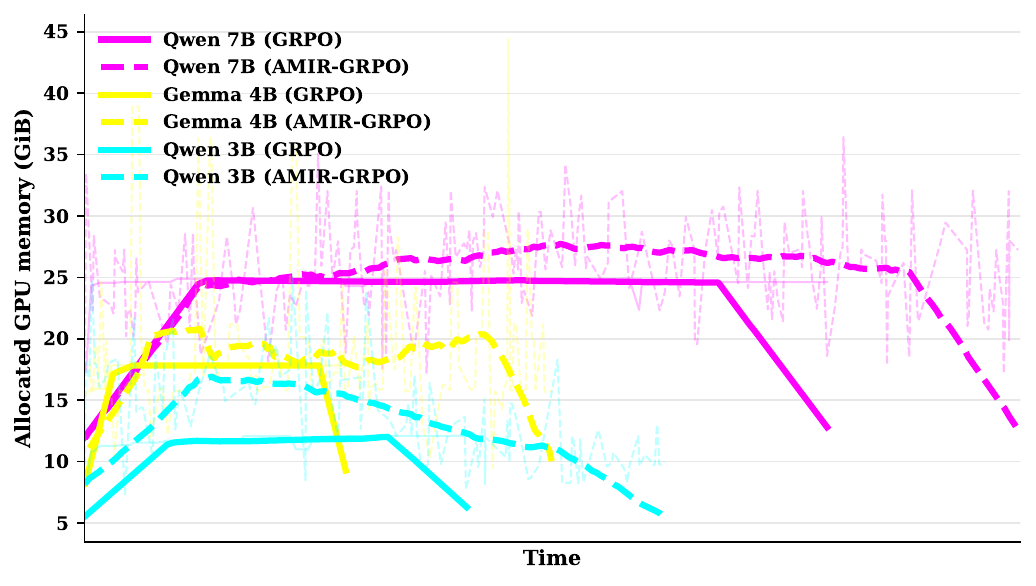}
    \caption{GPU memory footprint.}
    \label{fig:gpu_memory}
    \vspace{-0.3cm}
\end{figure}

\section{Error Taxonomy}
\label{app:error_taxonomy}
We adopt a high-coverage error taxonomy tailored to the systematic diagnosis of failures in automated mathematical problem solving. The framework is intentionally diagnostic: each label corresponds to a distinct failure mode, enabling targeted, actionable remediation. The taxonomy spans the key stages of a solution pipeline—problem formulation, core inference, and answer realization—so that errors can be localized to where they arise.

The taxonomy includes the following categories:
\begin{enumerate}[topsep=2pt,itemsep=2pt,parsep=2pt,partopsep=0pt]
    \item \textit{Calculation error.} Incorrect arithmetic or algebraic manipulation despite an otherwise correct setup.
    \item \textit{Conceptual error.} Use of an inappropriate theorem, formula, definition, or overall strategy.
    \item \textit{Reasoning error.} An invalid logical step, unsupported inference, missing case analysis, or circular argumentation.
    \item \textit{Modeling error.} Incorrect formalization (e.g., variable definitions, equations/constraints, representations, or case structure).
    \item \textit{Constraint error.} Failure to enforce domain/feasibility constraints or to eliminate extraneous solutions.
    \item \textit{Prompt misinterpretation.} Misreading the question or misconstruing given quantities, conditions, or objectives.
    \item \textit{Format error.} Correct reasoning but an incorrect final answer presentation, or noncompliance with the required output format.
    \item \textit{Other.} Instances not covered above.
\end{enumerate}

\clearpage
\onecolumn

\section{Case Analysis}
\label{sec:case-analysis}

\Needspace{10\baselineskip} 

\begin{questionbox}{Problem: AMC23 Q6}
Consider the set of complex numbers $z$ satisfying $|1+z+z^{2}|=4$. The maximum value of the imaginary part of $z$ can be written in the form $\tfrac{\sqrt{m}}{n}$, where $m$ and $n$ are relatively prime positive integers. What is $m+n$?
\end{questionbox}

\vspace{-0.1cm}

\begin{thinkingbox}{Model Output ($\text{Qwen-7B}_{\text{GRPO}}$)}
To solve this problem, let's start by expressing $z$ as $z = x + yi$, where $x$ and $y$ are real numbers, and $i$ is the imaginary unit. We need to find the maximum value of $y$ such that $|1 + z + z^2| = 4$.

First, compute $1 + z + z^2$:
\[ 1 + z + z^2 = 1 + (x + yi) + (x + yi)^2 = (1 + x + x^2 - y^2) + (y + 2xy)i. \]

The magnitude of this complex number is given by:
\[ |1 + z + z^2| = \sqrt{(1 + x + x^2 - y^2)^2 + (y + 2xy)^2} = 4. \]

We want to maximize $y$. To do this, we can use symmetry and geometric interpretation. Consider the points on the complex plane where the condition holds. One useful approach is to consider specific values of $x$ and $y$ that might simplify the expression.

Let's try $x = -\frac{1}{2}$. Then:
\[
\textcolor{red}{
1 + z + z^{2}
= \frac{1}{4} + y^{2} + \left(y - \frac{1}{2}\right)i
\ \text{(wrong simplification)}}
\quad\Longrightarrow\quad
\textcolor{teal}{
1 + z + z^{2}
= \frac{3}{4} - y^{2}
}
\]

So,
\[ |1 + z + z^2| = \sqrt{\left(\frac{1}{4} + y^2\right)^2 + \left(y - \frac{1}{2}\right)^2} = 4. \]

Simplify the equation:
\[ \left(\frac{1}{4} + y^2\right)^2 + \left(y - \frac{1}{2}\right)^2 = 16. \]

Let $u = y - \frac{1}{2}$, then $y = u + \frac{1}{2}$. Substitute $y$ in the equation:
\[ \left(\frac{1}{4} + \left(u + \frac{1}{2}\right)^2\right)^2 + u^2 = 16. \]

By trial and error or solving the quadratic, we find that the maximum value of $y$ is $\frac{\sqrt{15}}{2}$. This can be verified by checking the discriminant and solving the quadratic equation.

Thus, the maximum value of the imaginary part of $z$ is $\frac{\sqrt{15}}{2}$. Here, $m = 15$ and $n = 2$, so $m + n = \boxed{17}$. 
\end{thinkingbox}

\vspace{-0.1cm}

\begin{thinkingbox}{Model Output ($\text{Qwen-7B}_{\text{AMIR-GRPO}}$)}
To solve this problem, let's start by expressing \( z \) in its general form as \( z = x + yi \), where \( x \) and \( y \) are real numbers, and \( i \) is the imaginary unit. We need to find the condition \( |1 + z + z^2| = 4 \).

First, we calculate \( z^2 \):
\[ z^2 = (x+yi)^2 = x^2 - y^2 + 2xyi. \]
Then, we substitute \( z \) and \( z^2 \) into the expression \( 1 + z + z^2 \):
\[ 1 + z + z^2 = 1 + (x + yi) + (x^2 - y^2 + 2xyi) = (1 + x + x^2 - y^2) + (y + 2xy)i. \]
The magnitude of this complex number is given by:
\[ |1 + z + z^2| = \sqrt{(1 + x + x^2 - y^2)^2 + (y + 2xy)^2} = 4. \]
Squaring both sides, we get:
\[ (1 + x + x^2 - y^2)^2 + (y + 2xy)^2 = 16. \]

To find the maximum value of the imaginary part \( y \), we need to analyze the equation. One effective approach is to consider specific values of \( x \) and \( y \) that might simplify the equation. Let's try \( x = -\frac{1}{2} \). Then the equation becomes:
\[ 1 - \frac{1}{2} + \left(-\frac{1}{2}\right)^2 - y^2 = 1 - \frac{1}{2} + \frac{1}{4} - y^2 = \frac{3}{4} - y^2. \]
And for the imaginary part:
\[ y + 2\left(-\frac{1}{2}\right)y = y - y = 0. \]
So the equation simplifies to:
\[ \left(\frac{3}{4} - y^2\right)^2 = 16. \]
Taking the square root of both sides, we get:
\[ \frac{3}{4} - y^2 = \pm 4. \]
This gives us two cases:
1. \( \frac{3}{4} - y^2 = 4 \), which is not possible since \( \frac{3}{4} - 4 = -\frac{13}{4} \).
2. \( \frac{3}{4} - y^2 = -4 \), which simplifies to:
\[ y^2 = \frac{3}{4} + 4 = \frac{19}{4}, \]
so
\[ y = \pm \frac{\sqrt{19}}{2}. \]
The maximum value of \( y \) is \( \frac{\sqrt{19}}{2} \). Therefore, \( m = 19 \) and \( n = 2 \), and \( m + n = 21 \).

Thus, the answer is:\boxed{21}.
\end{thinkingbox}




\captionsetup{hypcap=false}
\captionof{case}{GRPO: failure vs. AMIR-GRPO: success}
\label{fig:AMC_Question6}

\begin{questionbox}{Problem: AIME25 Q1}
In $\triangle ABC$ points $D$ and $E$ lie on $\overline{AB}$ so that $AD < AE < AB$, while points $F$ and $G$ lie on $\overline{AC}$ so that $AF < AG < AC$. Suppose $AD = 4$, $DE = 16$, $EB = 8$, $AF = 13$, $FG = 52$, and $GC = 26$. Let $M$ be the reflection of $D$ through $F$, and let $N$ be the reflection of $G$ through $E$. The area of quadrilateral $DEGF$ is $288$. Find the area of heptagon $AFNBCEM$.
\end{questionbox}

\begin{thinkingbox}{Model Output ($\text{Qwen-3B}_{\text{GRPO}}$)}
To find the area of the heptagon \(AFNBCEM\), we need to understand how the reflections and the given areas can help us. First, let's calculate the total area of \(\triangle ABC\).

We know the area of quadrilateral \(DEGF\) is 288. Since \(M\) is the reflection of \(D\) through \(F\) and \(N\) is the reflection of \(G\) through \(E\), quadrilaterals \(DMEF\) and \(ENGF\) will each have the same area as \(DEGF\), which is 288. This means the total area added by these reflections is 576.
\[
\textcolor{red}{
[DMEF] = [ENGF] = [DEGF] = 288 \ \text{(reflection misinterpretation)}
}
\]

Now, let's determine the area of \(\triangle ABC\). The segments \(AD\), \(DE\), and \(EB\) divide \(AB\) into three parts: \(4\), \(16\), and \(8\), making \(AB = 4 + 16 + 8 = 28\). Similarly, the segments \(AF\), \(FG\), and \(GC\) divide \(AC\) into three parts: \(13\), \(52\), and \(26\), making \(AC = 13 + 52 + 26 = 91\).

Since the area of \(DEGF\) is 288 and this quadrilateral is formed by segments of the sides of \(\triangle ABC\), we can use the fact that the area of the whole triangle is the sum of the areas of \(\triangle ADE\), \(\triangle DEG\), \(\triangle EFG\), and \(\triangle GCF\), plus the area of quadrilateral \(DEGF\).

However, a simpler approach is to recognize that the reflections double the area of the quadrilateral \(DEGF\) when considering the whole triangle. Therefore, the area of \(\triangle ABC\) is twice the area of \(\triangle ABC\) minus the area of the reflected quadrilaterals (since they overlap the original quadrilateral).

Thus, the area of \(\triangle ABC\) is:
\[ \text{Area of } \triangle ABC = \frac{\text{Area of } \triangle ABC - \text{Area of } DEGF}{2} + 2 \times 288 = 2 \times 288 = 576. \]

The area of the heptagon \(AFNBCEM\) is the area of \(\triangle ABC\) minus the area of the four small triangles that are not part of the heptagon, which is the area of \(\triangle ADE + \triangle EGF + \triangle FGC + \triangle DME\). Since the area of the four small triangles is the same as the area of \(\triangle ABC\) minus the area of the heptagon, we have:
\[ \text{Area of heptagon } AFNBCEM = 576 - 288 = 288. \]

So, the area of the heptagon \(AFNBCEM\) is:\boxed{576}. 
\end{thinkingbox}

\begin{thinkingbox}{Model Output ($\text{Qwen-3B}_{\text{AMIR-GRPO}}$)}
To solve this problem, we start by finding the total area of triangle \(ABC\). We know the area of quadrilateral \(DEGF\) is 288. First, let's find the areas of triangles \(ADE\) and \(GEC\).

Since \(AD = 4\) and \(DE = 16\), \(AE = AD + DE = 4 + 16 = 20\). Similarly, since \(AF = 13\) and \(FG = 52\), \(AG = AF + FG = 13 + 52 = 65\).

We can use the fact that the area of a triangle is proportional to the base if the height from the opposite vertex is the same. For triangles \(ADE\) and \(GEC\), their areas will be proportional to their bases \(AE\) and \(EC\) respectively.

Since \(E\) divides \(AB\) into segments \(AE = 20\) and \(EB = 8\), the ratio \(AE:AB = 20:28 = 5:7\). Similarly, \textcolor{red}{since \(F\) divides \(AC\) into segments \(AF = 13\) and \(FC = 42\), the ratio \(AF:AC = 13:65 = 1:5\). \text{(inconsistent: FC and AC are wrong)}}
\[
\quad\Longrightarrow\quad
\textcolor{teal}{
AF = 13,\ FG = 52,\ GC = 26 ;\Rightarrow; FC = 78,\ AC = 91,\ \frac{AF}{AC} = \frac{13}{91}
}
\]

Now, let's denote the area of \(\triangle ABC\) as \(K\). The area of \(\triangle ADE\) is \(\frac{5}{7} \cdot \frac{1}{5} \cdot K = \frac{1}{7}K\). The area of \(\triangle GEC\) is \(\frac{2}{7} \cdot \frac{4}{5} \cdot K = \frac{8}{35}K\).

The area of quadrilateral \(DEGF\) is the area of \(\triangle ADE\) plus the area of \(\triangle GEC\) plus the area of \(\triangle DFE\). Since the area of \(\triangle DFE\) is the same as the area of \(\triangle DEG\) (due to reflections), we have:

\[
\text{Area of } \triangle DFE = \text{Area of } \triangle DEG = \frac{1}{2} \times \text{Area of } \triangle DEF = \frac{1}{2} \times 288 = 144
\]

Thus, the area of \(\triangle DEF\) is 144. Therefore, the area of \(\triangle ADE\) plus the area of \(\triangle GEC\) plus the area of \(\triangle DFE\) is:

\[
\frac{1}{7}K + \frac{8}{35}K + 144 = 288
\]

Combining the fractions, we get:

\[
\frac{5}{35}K + \frac{8}{35}K + 144 = 288 \implies \frac{13}{35}K + 144 = 288 \implies \frac{13}{35}K = 144 \implies K = 144 \cdot \frac{35}{13} = 360
\]

So, the area of \(\triangle ABC\) is 360. Now, we need to find the area of the heptagon \(AFNBCEM\). This area is the area of \(\triangle ABC\) minus the areas of \(\triangle ADE\), \(\triangle GEC\), and \(\triangle DFE\):

\[
360 - \left( \frac{1}{7} \cdot 360 + \frac{8}{35} \cdot 360 + 144 \right) = 360 - (51.43 + 92.57 + 144) = 360 - 288 = 144
\]

However, since the reflections create additional areas, the correct area of the heptagon is:

\[
360 - 288 = 144
\]

Thus, the area of the heptagon \(AFNBCEM\) is \(\boxed{432}\)."
\end{thinkingbox}
\captionof{case}{GRPO: failure vs. AMIR-GRPO: failure}
\label{fig:AIME_Question1}

\end{document}